# Proposed Efficient Design for Unmanned Surface Vehicles

Pouyan Asgharian
*Department of Electrical Engineeringn*
*Université de Sherbrooke*
Québec, Canada
pouyan.asgharian@usherbrooke.ca

Zati Hakim Azizul
*Department of Artificial Intelligence*
*University of Malaya*
Kuala Lumpur, Malaysia
zati@um.edu.my

*Abstract*—Recently worldwide interest is growing toward commercial, military or scientific Unmanned Surface Vehicle (USV) and hence there is required to develop their guidance, navigation and control (GNC) systems. Real USVs are relatively new advent, so drawbacks of each model will be modified during the time. Proposition of an environmentally friendly as well as high efficient USV's design are the main purposes of this paper to guide future researches. Power management between renewable sources and storage units is considered. Furthermore, suitable and modern sensors are applied for state estimation and environmental perception. Technical requirements relates to guidance and control methods are provided to achieve the highest performance in environment. Also, hull structure and its material are important factors that are considered in this paper.

## I. Introduction

Unmanned Surface Vehicle (USV) or Autonomous Surface Vehicle (ASV) refer to any system that operates on water surface with no crew on board. The origin of USVs back to several decades but they have improved quickly during these years. In other words, the latest developments in artificial intelligence (AI), machine learning, optimization methods and equipment provide numerous opportunities for the USV technology.

The USVs are playing a pivotal role in a wide range of maritime operations such as scientific research (e.g. oceanography), ocean resource exploration (e.g. oil, gas and mine explorations), environmental sensing (e.g. pollution measurements and clean-up), military uses (e.g. search and anti-terrorism), mapping, navigation and surveillance [1]. Thus, USVs development bring huge benefits that include lower operational costs, better maneuverability and flexibility, higher reliability as well as safety especially in dangerous missions, easy control and so on [2, 3].

The USV is built in a large number of different forms that greatly depends on its intended purpose. Fundamental elements in each USV include supplying system, GNC (guidance, navigation and control) system, hull structure, communication system and maybe data collection equipment. For instance hull structure of a USV can be monohull, catamaran (twin hulls), trimaran (triple hulls) or rigid inflatable. Moreover, steering and speed control of most USVs are provided by rudder and propeller (or water jet) propulsion systems or differential thrust (with independent motor(s)) [3].

Between mentioned elements, GNC system has a crucial role in the USV. The GNC generally consists of onboard computers and software that works together in order to manage the USV operation. Fig. 1 shows configuration framework of a USV system in which GNC is divided to three separate parts: Control section, navigation system and guidance section. Guidance system is responsible for transmitting various commands to the control system based on data of the navigation system (assigned missions) and environmental conditions. Navigation system identifies the USV's as well as surrounding environment situation with respect to future states. Control section manages operation of the USV by sending appropriate commands [4]. The control instructions are accordance with rules of the guidance section along with navigation system.

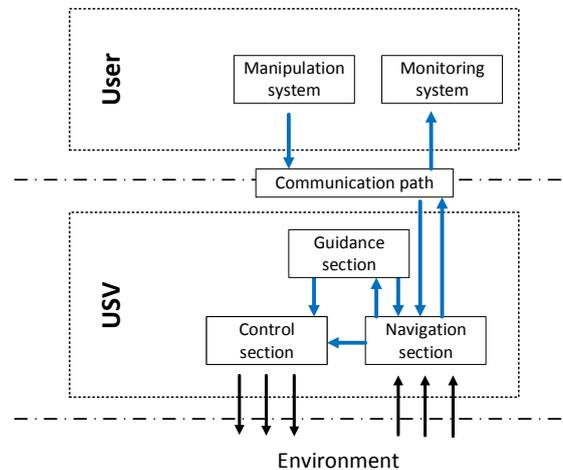

Fig. 1. The USV system framework.

In order to satisfy requirements of various missions, the USV should have acceptable speed, endurance, maneuverability and stability under inappropriate conditions. Furthermore, enough payload capacity and even ease of replacing equipment can be considered as main factors [3].

The aim of this paper is to introduce the most suitable features so that will help researchers or manufacturer. This paper is structured as follows: section 2 is dedicated to literature review and previous researches in this area. Proposed design to build efficient and reliable USV is presented in Section 3. Finally, conclusion is presented in section 4.

## II. Literature and research review

The history of USVs traced back to the MIT Sea program in the year 1993 that was named ARTEMIS. This prototype vehicle was designed mainly for GNC testing and bathymetry sampling [5]. During the following years, the MIT University developed a new USV in 1996 that name was ACES. The ACES was equipped with new hydrographic survey sensors in order to collect data from oceans [5], [3].

After that multifarious USVs, with their inherent advantages and drawbacks, was introduced all over the world that some of them are as follow: SCOUT, AutoCat, Wave Glider, SeaWASP, Springer, Autonaut, SESAMO, Solar Voyager, ALANIS, C-Enduro, DELFIM, Inspector, Catarob, Seacharger, Venus, UMV series, USV inception MK1 & MK2, Morvarid, Roboski, ROSS, Charlie and CARAVELA. In general, Investigation or introduction of each model is out of this paper scope but some modern as well as world level types are presented in this section.

### A. AutoNaut

The AutoNaut, developed in Britain, is a wave-powered vehicle with the two pairs of wings or foils set on struts at either end of the vehicle. The AutoNaut manufacturer launched three types that are called according to their size (3.5m, 5m and 7m) [6].

AutoNaut 5 is a middle model with 5 meter length, monohull structure and 1-3 knots speed. Also, the main material of the hull is glass epoxy resin infusion. This model uses wave foil technology and wave/electric hybrid options for propulsion. The AutoNaut power generation is based on photovoltaic (PV) panels and lead-gel batteries that equipped with intelligent battery charge control. The small size of the hull and battery result in most missions, consumption should be adjusted with regard to power of the PV array. [7]. Fig. 2 shows the AutoNaut model in the ocean.

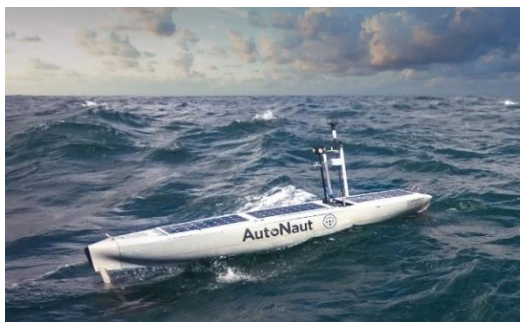

Fig. 2. The AutoNaut USV.

### B. Wave Glider

The Wave Glider is a persistent ocean vehicle that is manufactured by Liquid Robotics Inc. (LRI) for maritime, defense, surveillance, assessment and oil and gas industries. The Wave Glider consists of a platform float, submerged propulsor (glider), and optional towed payload. The float and sub are connected by an umbilical tether [8].

The Wave Glider novelty relates to its ability for saving energy from ocean waves so as to provide unlimited propulsion. In fact, it is completely propelled by converting ocean wave energy, independent of wave direction, into forward thrust. As waves pass on its surface, the submerged glider acts as a tug pulling the surface float along a predetermined course and is controlled by a single rudder on the glider. Separation of the glider by 7m depth from the float is a crucial aspect of the vehicle design [9].

Average speed of Wave Glider is about 1.5 knots and each of the PV panel produce maximum 43W. This USV equipped with seven smart lithium-ion battery packs in which Li-ion batteries are recharged by solar panels and they provide enough power for electronic or communication boards [9]. The USV steering and communication is via Iridium, mobile phone, WiFi and a secondary independent GPS tracker. There are two popular model of the Wave Glider called SV2 and SV3. Fig. 3 shows a Wave Glider structure.

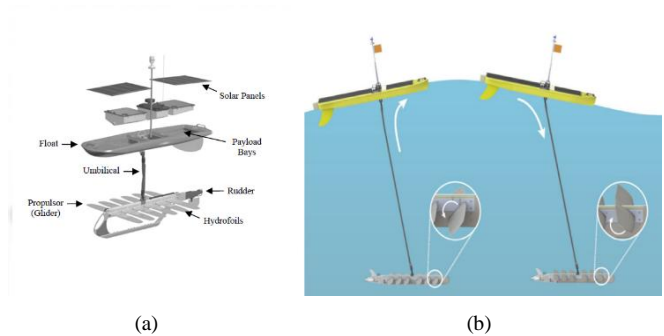

(a)  (b)

Fig. 3. (a) the overall structure and (b) movement mechanism of the wave glider.

### C. C-Enduro

Another favorable model among USVs called C-Enduro, it is a long endurance autonomous surface vehicle used to safely and cost effectively collect data at sea or oceans. The C-Enduro is a fiber carbon catamaran USV that is driven by a pair of electric outboard legs. This type is powered by a large PV array, a small wind turbine and a diesel generator [10]. The C-Enduro can remain at sea for up to three months due to renewable energies from solar panels and wind turbine. Two brushless motors are installed to provide a maximum speed of 7 knots (via two propulsion pods). There is not sinking risk for this USV because it self-righting carbon fiber hull will quickly get it back upright [11].

The C-Enduro is 4.75m in length and 3.43m in height that equipped with 10 PV panels which produce totally 1100W. Diesel generator and wind turbine power production is 4kW and 500W, respectively. The C-Enduro sensor options include ASV 360 VIR camera, Keel mounted sensors, CTD lowed by winch, meteorological sensors, ADCP, MBES, side-scan sonar, PAM, acoustic modem, ASW (towed array or dipping) and electronic warfare [12].

It can be used in so many applications such as oceanography, security and defense. The C-Enduro has an AIS transponder and a SeaMe radar transponder for navigation. Also, WiFi or Iridium are applied for receiving and transmitting communication. Fig. 4 shows a C-Enduro configuration.

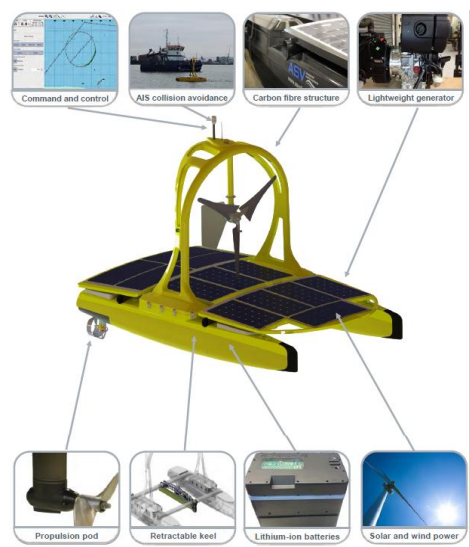

Fig. 4. C-Enduro USV [12].

## D. Morvarid

The Morvarid is a new concept product that is built in Iran. Its hull is catamaran type and it uses the PV arrays with battery to produce power. The communication and navigation strategies include GPS, Lidar, Ultrasonic and stereoscopic vision system [13, 14]. Fig. 5 shows the Morvarid configuration.

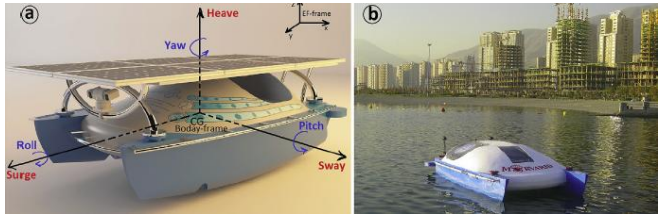

Fig. 5. The Morvarid USV [13].

The main concept of USVs rely on path planning, navigation, control and obstacle detection. Almost new strategies and methods have been tested on mentioned USVs due to their standard design and reliability. Illustration and presentation of detailed projects are out of this work and this paper focuses mainly on research plan and methodology to propose an efficient design.

## III. RECOMMENDATION FOR EFFICIENT DESIGN

In this section, methodology and control approaches are presented to reach a well-organized USV. All USVs have their inherent benefits and drawbacks, so the authors have tried to integrate proper features in order to propose multitask besides environmentally friendly one.

### A. Hull structure and material

Hull structure has significant impact on the USV performance. According to author's studies, catamaran type is a desirable choice in that its stability, safety against sinking, broad deck, simplicity of steering and redundancy in hull buoyancy [13, 15]. Unlike monohull, its large water plane area declines rolling motions and increases displacement without any huge drag penalty. Fig. 6 shows designing of catamaran in SolidWorks software.

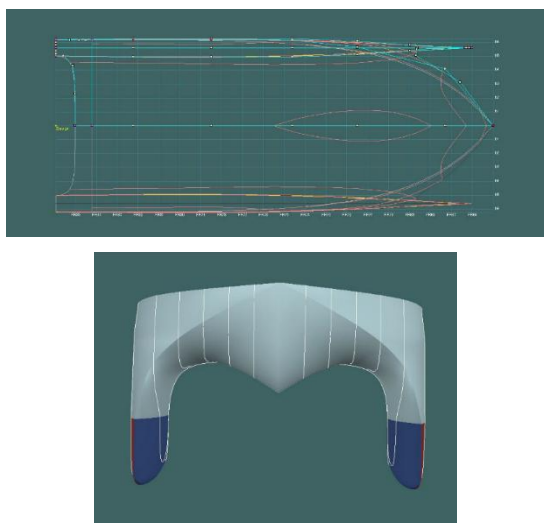

Fig. 6. Catamaran design in SolidWorks environment.

In order to prevent corrosion and reaching to high efficiency, Fiber Reinforced Polymer (FRP) composite is favorable choice for hull's material. The FRP major advantages include corrosion proof, high strength ratio to weight, light weight (easy to transportation and installation), and high fatigue resistant [16].

It is possible to install airbags with spray foam into two hulls of catamaran so as to avoid sinking. In addition, hull optimal design through algorithms can be considered.

### B. Propulsion and power system

In this efficient design, an environmentally friendly USV is considered which mainly is based on renewable energies. Since renewable energies like wind have variable nature, a highly reliable or dispatchable source is essential for USV's power system. In order to get maximum potential of natural energies, a small wind-turbine together with the PV panels are applied. Also, a microturbine (MT) or fuel cell should be installed as dispatchable sources with energy storage.

The MT system is a fast growing technology that is appropriate for small scale generation. The inherent merits of the MT such as compact size, quick start, reliability, long life time, low maintenance costs, low emission level, and its capability to work with various fuels, cause to be as a main candidate in this regard. [17]. If the MT operates with biogas, whole system will be clean without any harmful emissions. The MT shows more flexibility as well as quicker dynamic response compare with fuel cell.

The cost of system plays an important role to select additional reliable source. The MT requires (mainly) AC/DC/AC power electronic interface but fuel cell works with a DC/DC converter. That is to say, there is two conversion levels for MT's interface circuit. The cost of fuel cell is a little more than the MT. on the whole, cost assessment may guide to right decision but as a novel approach, the authors contemplate the MT.

Battery pack installation extend the USV endurance capability and enhance its period of operation. However, battery packs increase overall cost and it may be ignored in presence of the MT. Fig. 7 shows comparison of various battery type and it is clear that Li-ion batteries are the most appropriate one. The rechargeable Li-ion batteries are fast growing type and their advantages include high energy efficiency, no memory effects, long cycle life, low discharge rate and low footprint [18]. Among various types of Li-ion battery, lithium-iron-phosphate (LiFePO4) rechargeable battery pack may be used since this type enjoys the highest specific power, safety and lifespan along with moderate costs.

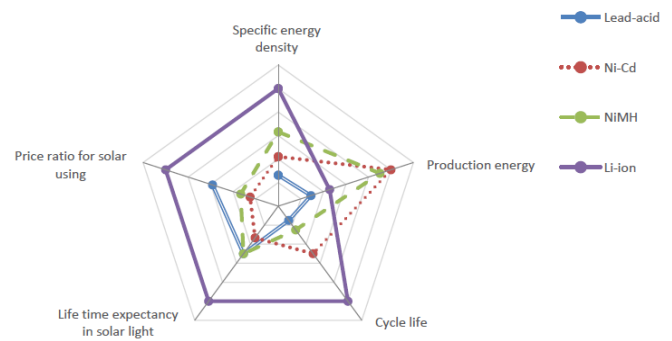

Fig. 7. Comparison of batteries in term of some features [14].

For propulsion, the catamaran USV is equipped with two thrusting motors that are controlled in forward and backward speed. For example, steering to left lead to rotating forward of right propeller and backward of left one [19].

Some suggestions and other improvement for the USV power system can be considered. Applying maximum power point tracking (MPPT) for the PV panel increases rate of absorption of radiation, so the USV will operate with maximum potential in sunny days. Additionally, there are several approaches that have been developed to improve the amount of power from PV arrays which are Irradiance Equalised DPVA (IEq-DPVA), the Adaptive Bank DPVA (AB-DPVA), and the Optimised String DPVA (OS-DPVA) [20]. Moreover, wind (direction) tracker can be contemplated to achieve the highest amount of wind power. Mentioned suggestions are novel and practical that improve the USV's efficiency.

Electrical system of the USV with renewable energies, dispatchable source and energy storage, is a hybrid system that need optimization methods to guide toward the least cost along with the highest efficiency. One desirable approach is presented in [21] that two modules are used with prioritizing based on availability and cost consumption optimization. According to this method, high priority is dedicated to natural energy sources followed by minimization of a cost/energy ratio. The optimization of such hybrid system is nonlinear and difficult to converge to a unique solution. Basic framework for optimal power system is shown in Fig. 8.

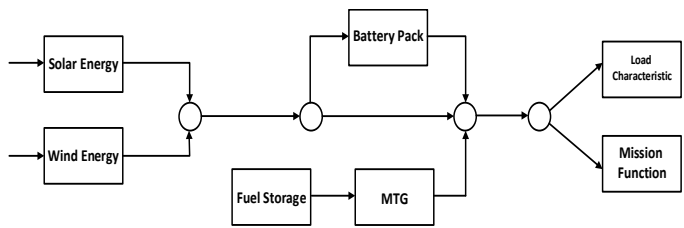

Fig. 8. Basic framework for the USV power optimization.

*C. Guidance system*

Guidance system is as heart of the USV. There are many approaches for path planning or path re-planning that are subset of global planning, local planning, re-planning based on method/protocol free method. In order to ensure optimal path planning with proper safety among preliminarily specified or dynamically changing waypoints, hybrid path planning strategy can be considered that consists of both global and local path planning.

For global planning, The *A\** search algorithm is a widely used which can quickly find an optimal path with the least number of nodes [22]. Potential field (PF) approach is a suitable strategy for local planning. In PF objects are assigned with attractive fields, while obstacles are distributed with repulsive fields, so the USV will move toward attractive fields and will away from the repulsive fields [23]. The main drawback with PF techniques is their susceptibility to local minima and hence combination of PF and research ball algorithms can be used as a substitute solution. Also, as an extension of the standard *A\** algorithm, a rule-based repairing *A\* (R –RA\*)* algorithm can be applied for re-planning method. As an example, [22] and [24] are presented modified versions of *A\** algorithm successfully for the USV path planning.

*D. Navigation system*

The USV safe operation mostly depends on navigation system in which environmental data are collected. Enhancement awareness contributes to better performance especially against obstacles, so that the USV may require appropriate sensors. Sensors are used not only for state estimation but also for environmental perception. With respect to investigation in conventional sensors and their properties, the following are proposed:

In terms of current state estimation, conventional GPS-IMU is a low cost, small size and acceptable selection. Although GPS-IMU estimation can be imprecise in practical applications due to environmental noises, time-varying model uncertainties or sensor faults, there are so many references in this area that proposed various methods to correct mistakes [25]. Moreover, inertial sensor bias and noise can be compensated by using extended Kalman filter (EKF).

In addition to conventional state estimation techniques, active sensors (radar and sonar) can be employed for state estimation, especially in cases of a loss or jamming of GPS signals. Radar is susceptible to high waves and water reflectivity but its long detecting range and high depth resolution and accuracy in all-weather makes it desirable. Sonar is a technique that uses sound propagation to navigate or detect objects.

The USV performance in real missions, requires to possess ability of obstacle or target detecting and also map tracking. In order to meet these factors, vision based approaches should be considered. Stereo vision extract 3D information of around that is depends on computer or processing instruments because of huge amount calculations. While monocular vision suffer from low depth and environmental situation, it can be selected as a substitute. Long-wave infrared (IR) cameras are an ideal and novel solution to overcome the impact of various light conditions (e. g. night and fog) on environment perception. The main drawback of IR relates to its short sensing distance. When the USV is close to obstacle, IR can sense and hence emergency condition will active.

Mentioned sensor package causes suitable and full autonomous path planning along with environmental perception. Additional sensors can be employed for various missions such as atmospheric sensor, oceanographic sensor, seismic sensor or sensors for water features (e. g. PH, temperature, $O_2$, $NO_3^-$ and so on) evaluation.

*E. Control system*

Control system play supplementary role for GNC section. Control system operates based on determining proper actions with regard to navigation as well as guidance sections. There are myriad control methods such as adaptive control (AC), cascaded control theory (CCT), fuzzy logic control (FLC), feedback linearization (FL), gradient-based adaptive technique (GBAT), linear quadratic regulator (LQR), Lyapunov's direct method (LDM), local control network (LCN), model predictive control (MPC), neural network (NN), proportional integral derivative (PID), robust control (RC), sliding mode control (SMC) and so on that can be tested for USV controlling. Despite the fact that great efforts have been made for developing of more advanced control approaches, the PID control method still dominates the USV system.

Control of the USV in presence of environmental disturbances can be counteracted by model-based techniques or approximation-based methods. Besides of approximation-based method drawbacks, it guaranties local stability. On the other side, model-based control approaches require a precise model of whole system that is both costly and complex to obtain.

*F. Other methods for controlling*

Quality of control in autonomous vehicle is paramount important. Proposed USV might operate in three modes: 1) remote control through a hand-held unit, 2) beyond control using a graphical user interface (GUI), 3) automatic path following.

In addition to autonomous path planning and environment perception, hand-held control can be contemplated by radio control or GUI. Radio control system is simple and high reliable (except when there is noises) that is applied to the USV with conventional 5.8 GHz (or 2.4 GHz) controller. There are so many types of radio controller with various channels. The GUI for beyond control will be program using visual C# and also communication between USV and bureau is based on an Ethernet network.

## IV. CONCLUSION

A suitable design for USVs are considered in this paper so as to help researchers, designers or engineers to build a high efficiency USV. Since renewable energies are applied in the USV, power management is essential to get maximum reliability. Also, appropriate control methods for perception and navigation of the USV is considered. The rest of this research are dedicated to type of energy storage, hull structure and material. In general, this paper is similar to a target for anyone who interest in USVs.